\documentclass[11pt]{article}

\usepackage{amsmath}
\usepackage{theorem}
\usepackage{times}
\usepackage{bm}
\usepackage{natbib}
\usepackage[plain,noend]{algorithm2e}

\usepackage{graphicx}
\usepackage{color}
\usepackage{xcolor}

\makeatletter
\renewcommand{\algocf@captiontext}[2]{\quad #1\algocf@typo. \AlCapFnt{}#2} 
\def\@algocf@capt@plain{top}
\renewcommand{\algocf@makecaption}[2]{%
  \addtolength{\hsize}{\algomargin}%
  \sbox\@tempboxa{\algocf@captiontext{#1}{#2}}%
  \ifdim\wd\@tempboxa >\hsize
    \hskip .5\algomargin%
    \parbox[t]{\hsize}{\algocf@captiontext{#1}{#2}}
  \else%
    \global\@minipagefalse%
    \hbox to\hsize{\box\@tempboxa}
  \fi%
  \addtolength{\hsize}{-\algomargin}%
}
\makeatother

\def\v{{\varepsilon}}
\newcommand{\ve}[1]{\bm{{#1}}}
\newcommand{\vesub}[2]{\bm{{#1}}_{#2}}
\newcommand{\vesup}[2]{\bm{{#1}}^{#2}}
\newcommand{\vess}[3]{\bm{{#1}}_{#2}^{#3}}
\newcommand{\hve}[1]{\hat{\ve{#1}}}

\newcommand{\tve}[1]{\tilde{\ve{#1}}}
\newcommand{\tvesub}[2]{\tilde{\ve{#1}}_{#2}}
\newcommand{\tvesup}[2]{\tilde{\ve{#1}}^{#2}}
\newcommand{\tvess}[3]{\tilde{\ve{#1}}_{#2}^{#3}}

\newtheorem{theorem}{Theorem}[section]
\newtheorem{remark}{Remark}[section]
\newtheorem{lemma}{Lemma}[section]

\usepackage{geometry}                
\usepackage{graphicx}
\usepackage{amssymb}
\usepackage{epstopdf}
\title{Distributed sequential federated learning}
\author{Z. F. Wang\footnote{zfw@ustc.edu.cn}, 
\space 
X. Y. Zhang\footnote{ xinyu@amss.ac.cn}\\
International Institute of Finance, The School of Management,\\ University of Science and Technology of China, China \\ \\
and\\ \\
Y-c I. Chang\footnote{ycchang@sinica.edu.tw}\\
Institute of Statistical Science, \\
Academia Sinica, Taipei 11529, Taiwan 
}

\begin{document}
\maketitle

\begin{abstract}
The analysis of data stored in multiple sites has become more popular, raising new concerns about
the security of data storage and communication.  Federated learning, which does not require centralizing data, is a common approach to preventing heavy data transportation, securing valued data, and protecting personal information protection. Therefore, determining how to aggregate the information obtained from the analysis of data in separate local sites has become an important statistical issue.  The commonly used averaging methods may not be suitable due to data nonhomogeneity and 
incomparable results among individual sites, and applying them may result in the loss of information obtained from the individual analyses.
Using a sequential method in federated learning with distributed computing can facilitate the integration  and accelerate the analysis process. We develop a data-driven method for efficiently and effectively aggregating valued information by analyzing local data without encountering  potential issues such as information security and heavy transportation due to data communication.  In addition, the proposed method can preserve the properties of classical sequential adaptive design, such as data-driven sample size and estimation precision when applied to generalized linear models. We use numerical studies of simulated data and an application to COVID-19 data collected from 32 hospitals in Mexico, to illustrate the proposed method.
\end{abstract}

\noindent{Keywords:}
Adaptive sampling; Data Communication; COVID-19; Random average; Sequential sampling

\section{Introduction}

When data are independently collected in many separate sites,  data transportation and machine/data communication are essential issues. In addition to efficiency issues in data transportation and/or communication, information security  has become even more crucial \citep{damiani2015distributed,huang2020}.
If maintaining a centralized data warehouse is impractical due to cost, security and/or other reasons, federated learning is a feasible option that enables data scientists to manage and utilize all data for their analysis tasks. The key to this type of method is avoiding obstacles due to data communication and transportation. However, the discussions of federated learning in the machine learning literature, focus more on the perspectives of information technicality \citep{yan2013,Jordan2018,li2018,Goetz2019,Li2020,zhou2020}, and there is a lack of consideration of statistical issues.  

The common features of these methods are the analysis of data separately at each site, and the integration of the results from individual sites together in a convenient way. 
However, the nonhomogeneity among individual site data may pose a challenge for result integration via usual methods such as weighted average and statistical robust methods that bypass this issue by treating the results obtained from the separated sites as random observations \citep{McMahan2017}. Although some methods can deal with outliers, they are not efficient and effective in terms of sample size and will usually not retain all the information from the individual analyses. 
The issues of legitimately and effectively merging the local results from individual data sites and the corresponding statistical properties urgently require further
\citep{Dwork2011, Carlini2019, zhou2020}.
Thus, developing a method that addresses the original issues raised in the common federated learning discussion and can obtain better and more precise information of interest is the aim of this research.

In this paper, we propose a novel sequential method that takes advantage of both distributed computing and sequential sampling.
We adopt the idea of sequential analysis within each local site as a means of efficiently and effectively using the data in each site and then study how to integrate the results from the sepqrate analyses based on their formation contents in terms of both quantity and quality, instead of using the common weighted average which mostly depends on the prefixed sample sizes of the local sites. 
Because the data sets were collected before-hand, sampling is not an obstacle as it is in conventional clinical trials where recruiting new subjects into a study is rather complicated \citep{Woodroofe1982, Bartroff2013, Whitehead2014}.  We apply the proposed method to generalized linear models  and use logistic regression models to illustarte the details of our methods. 
We focus on classification problems, present the numerical results using simulated data and a real data set including data collected in 32 different hospitals in Mexico, and then summarize our findings.

\section{Federate sequential learning}\label{two-side} 
We apply the proposed method to the generalized linear model \citep[GLM, ][]{Nelder1972, McCullagh1989}, including logistic regression models, to illustrate the concept.

Suppose there are $M$ separate data sites, and $(Y, \vesub{X}{})$ are the response and covariate variables, respectively.
Then, for site $j$, with sample size $n_i$, $(Y_{ij}, \vesub{X}{ij})$, $j=1, \cdots, M$, $i=1, \cdots, n_j$ denotes its $i$th pair of observations.
Assume that the data of each individual site satisfy a generalized linear model with a link function $\mu$ such that
$ E(Y_j | \vesub{X}{j}) = \mu(\vess{X}{j}{\top}\vesub{\beta}{j})$ and
$ var(Y_j | \vesub{X}{j}) =\nu(\vess{X}{j}{\top}\vesub{\beta}{j})>0$,
where $\vesub{\beta}{j}$ $j=1, \cdots, M$, is an unknown length $p_j$ parameter vector. 
Define  $\ve{U}$ as a $p_0$ vector of common variables of interest within all sites.
Let $\vesub{V}{j}$ be a $p_j-p_0$ vector of other variables in site $j$. Note that the length of $\vesub{V}{j}$ may vary among sites.
Define $\vesub{X}{j}=(\vesup{U}{\top}, \vess{V}{j}{\top})^\top$, then with $\vesub{\beta}{j}=(\vesup{\theta}{\top}, \vess{\eta}{j}{\top})^\top$, we have that 
\begin{align}
& E(Y_j | \vesub{X}{j}) = \mu(\vesup{\theta}{\top}\ve{U}+\vess{V}{j}{\top}\vesub{\eta}{j}),\label{meanfun}\\
& var(Y_j | \vesub{X}{j}) =\nu(\vesup{\theta}{\top}\ve{U}+\vess{V}{j}{\top}\vesub{\eta}{j})>0.\label{varfun}
\end{align}

\begin{remark}
The generalized linear model family includes many popular models, where logistic regression models are commonly used in many fields. In these models, equations \eqref{meanfun} and \eqref{varfun}, implies that for  $j=1,\cdots, M$,
\begin{align}\label{model1}
&\mu(\vesup{\theta}{\top}\ve{U}+\vess{V}{j}{\top}\vesub{\eta}{j})=\frac{\exp{\vesup{\theta}{\top}\ve{U}+\vess{V}{j}{\top}\vesub{\eta}{j})}}{1+\exp{(\vesup{\theta}{\top}\ve{U}+\vess{V}{j}{\top}\vesub{\eta}{j})}}, \nonumber\\
&\nu(\vesup{\theta}{\top}\ve{U}+\vess{V}{j}{\top}\vesub{\eta}{j}) = \mu(\vesup{\theta}{\top}\ve{U}+\vess{V}{j}{\top}\vesub{\eta}{j})(1-\mu(\vesup{\theta}{\top}\ve{U}+\vess{V}{j}{\top}\vesub{\eta}{j})).
\end{align}
For a linear regression model, \eqref{meanfun} and \eqref{varfun} are
\begin{align}\label{model2}
& \mu(\vesup{\theta}{\top}\ve{U}+\vess{V}{j}{\top}\vesub{\eta}{j}) = \vesup{\theta}{\top}\ve{U}+\vess{V}{j}{\top}\vesub{\eta}{j},\nonumber\\
& \nu(\vesup{\theta}{\top}\ve{U}+\vess{V}{j}{\top}\vesub{\eta}{j}) =\sigma_j^2>0.
\end{align}
\end{remark}
If the sample sizes of the individual procedures are prefixed, then there are many classical methods for combining the results of $M$ procedures. Among them, some voting schemes and the concepts of weighted and robust statistical methods are popular choices. Because these methods are based on prefixed sample sizes, the variability among the estimates from different sites can be high when there are large differences in the sample sizes between sites. It is difficult to reduce the variation, even by choosing the same sample sizes for all sites, when the data quality of individual pools varies, which is common in some applications such as the context of the coronavirus pandemic. On the other hand, if results from individual sites have the required asymptotic statistical properties, then it is possible to combine them effectively and efficiently. 

We use sequential procedures to retain both parameter estimation and prediction power. 
This is novel in sequential analysis, to the best of our knowledge.
For generalized linear models, we use confidence set estimation with a prescribed accuracy for the parameter of interest, say $\ve\theta$ for the estimation quality. 
The index for prediction may vary depending on the types of response variables.
For logistic regression models, in addition to the size of the confidence set, we control the prediction performances via the area under the receiver operation characteristic (ROC) curve, named AUC, while for linear regression models, we calculate the mean square error (MSE) and/or corresponding $R^2$ statistics and use them as  performance indexes. 

Although the sample size of each procedure is random due to the stopping criterion and there is a lack of discussion about combining the results of many random sample-sized procedures, prioritizing the required statistical properties, instead of the sample sizes, seems more reasonable for analyzing combined data sets.

Here, we study valid integration of the results of $M$ procedures such that the final result has the desired property as in a conventional sequential procedure, and use the fixed-size confidence set estimation of logistic regression and linear models as examples. 
Because we can independently conduct $M$ estimation procedures without a centralized data center, the proposed method retains the key features of federated learning, such as protecting the personal information of the subjects at each site and reducing the data communication requirements.  
We now describe the $j$th sequential procedure, one of the $M$ sequential procedures, and the method for integrating $M$ independent results below.

\begin{remark}
By runing $M$ estimation procedures separately, we can decrease the computational load of each computer and, in fact, accelerate the computing efficiency by using smaller sample sizes in the individual computing facilities of local sites, since all sites can run their own procedures simultaneously. Thus, the proposed method can also benefit from a distributed computing framework, and each local site can be free to maintain its own high-power high-cost computing facility. This feature of the distributed sequential methods will certainly promote the uses of the federated learning methods.
\end{remark}

\subsection{Sequential estimate with fixed-sized confidence set and reserved prediction precision}
To combine the results, and overcome the issue of the nonhomogeneity among local sites, we require the confidence estimation of each procedure to achieve a given size and a prescribed accuracy. Instead, we allow the sample size used in each procedure to be variable and random, depending on the sequential sampling strategy and the quality of local data. 
The idea of sequential analysis is an appropriate tool for these purposes. 
Because we conduct $M$ procedures separately and independently with their own local data pools, we will just specify one of them in detail, say the $j$th procedure, and then focus more on how to combine results.

Let $C_{jn} = \{(y_{ji},\vesub{x}{ji}), i=1,...,n\}$ be the recruited data subset of the original data set $\mathcal D$ at stage $n$. 
Then at this stage, we compute the maximum quasi-likelihood estimate (MQLE) for $\ve\beta$ in \eqref{meanfun} and \eqref{varfun}, and denote them as $\tvesub{\beta}{jn}=(\tvess{\theta}{jn}{\top}, \tvess{\eta}{jn}{\top})^\top$, which are the solutions to the estimating equation \citep{McCullagh1989}:
\begin{eqnarray}\label{gle}
    ln(\tvesub{\beta}{jn})\equiv\sum_{i=1}^n \dot{\mu}(\vess{X}{j}{\top}\vesub{\beta}{j}) w(\vess{X}{j}{\top}\vesub{\beta}{j}) [y_{ji}-\mu(\vesub{x}{ji}^{\top}\tvesub{\beta}{jn})]\vesub{x}{ji}=0,
\end{eqnarray}
where $\dot{\mu}(t)=d\mu(t)/dt$ is the first derivative of $\mu(t)$ and $w(t)=\nu^{-1}(t)$.
Let $\vesub{L}{j}$ be a $p_0\times p_j$ matrix that consists of the first $p_0$ rows of the diagonal matrix, $\text{diag}\{I_{j1}, \cdots,I_{jp_j}\}$,  with $I_{j1}=\cdots=I_{jp_0}=1$ and $I_{jk}=0, k=p_0+1,\cdots,p_j$.
Then, the estimates of the parameters of interest are $\tvesub{\theta}{jn}=\vesub{L}{j}\tvesub{\beta}{jn}$.
Assume the following conditions are satisfied.
\begin{description}
\item{\em (A1)} $\sup_{i}||\vesub{x}{i}||<\infty$ for all variable $\vesub{x}{i}$, 
and $E|\epsilon_i|^\zeta<\infty$ with some $\zeta>2$, where $\epsilon_i = Y_i - \mu(\vesub{x}{ji}^{\top}\vesub{\beta}{j0})$ is the error term. 
\item{\em (A2)} 
$\lim_{n\rightarrow \infty}\sum_{i=1}^n\vesub{x}{ji}\{\dot\mu(\vesub{x}{ji}^{\top}\vesub{\beta}{j0})^2/\nu(\vesub{x}{ji}^{\top}\vesub{\beta}{j0})\}\vess{x}{ji}{\top}/n=\Sigma_j$, where $\Sigma_j$ is a positive definite matrix and $\vesub{\beta}{j0}$ is the true value of $\vesub{\beta}{j}$.
\end{description}
Then MQLE $\tvesub{\beta}{jn}$ of the GLM is a strong consistent estimate of $\vesub{\beta}{j}$ and has the asymptotic normal property \citep{Chen1999, Chang1999}.
It follows that
\begin{align}\label{thetaNorm}
    \sqrt{n}(\tvesub{\theta}{jn}-\vesub{\theta}{0})\longrightarrow
    N(0,{\vesub{L}{j}}\Sigma_j^{-1} {\vesub{L}{j}})
    ~~~~\text{\rm in distribution as }n\rightarrow \infty,
\end{align}
 where $\vesub{\theta}{0}$ is the true value of the common parameters of interest, $\ve{\theta}$.

\noindent{\bf{Performance measure and its estimate}} Without loss of generality, here we let  $A_j$ be a prediction measure, and $\hat{A}_j$ be a strongly consistent estimator of it with variance estimate $v_{Aj}$. 
It is straightforward to show that $(\hat{A}_j-A_j)/\sqrt{v_{Aj}}$ converges to $N(0,1)$ in distribution, as $n$ goes to $\infty$.
Let $\hat{y}^n_{ji}=\mu(\vess{x}{ji}{\top}\tvesub{\beta}{jn})$ be the fitted values of $y_{ji}$ when using the data set $C_{jn}$.

Let $S_1=\{\hat{y}_{ji}: y_{ji}=1\}$ and $S_0=\{\hat{y}_{ji}: y_{ji}=0\}$, and $n_0$ and $n_1$ be the sizes of $S_0$ and $S_1$, respectively, 
Then for a logistic regression model described by \eqref{model1},  \cite{zhou2009statistical} shows that
\begin{align}
\widehat{AUC}_j = \frac{1}{n_0n_1}\sum_{v_1\in S_1}\sum_{v_2\in S_0} I(v_1 \geq v_2),
\end{align}
is an estimate of the AUC using the data set $C_{jn}$, where $I(\cdot)$ is an indicator function. 
Let $\v{Aj}$ be the variance of $\widehat{AUC}_j$. It follows that $(\widehat{AUC}_j-AUC_j)/\sqrt{v_{Aj}}$ converges in distribution  to $N(0,1)$ as $n$ tends to $\infty$, where $AUC_j$ is the true AUC under (\ref{model1}).

\noindent{\bf{Sequential estimation procedure}}
Let $C_{jn_0}$ be the initial data set with size $n_0$ of procedure $j$, and $a$ be the square root of the $1-\alpha$ quantile of a
$\chi^2_{p_0}$ chi-square distribution with $p_0$ degrees of freedom.  Let $\tilde{a}_j>0$, $j=1, \cdots, M$,  such that $\sum_{j=1}^M\tilde{a}_j^2=a^2.$ { Values of $\tilde{a}_j$ are taken based on the customer's choice. For example, usually  $\tilde{a}^2_j = a^2/M$. When one data site $j$ has a small data size, $\tilde{a}_j$ should be assigned a small value to avoid a sequential procedure that cannot stop even using all samples at the site. Different $\{\tilde{a}_j: j=1,\cdots, M\}$ values do not affect the statistical properties of the final parameter estimation, as long as the sequential procedures satisfy for all $M$ sites. }
Let $a_{p}$ be the $1-\alpha/2$ quantile of $N(0,1)$.
Then, for a given $d_1, d_2 >0$, a stopping rule {$\tilde{N}_j$} is defined as
\begin{align}\label{Ndef-var}
\tilde{N}_j=N_{d_1, d_2}\equiv\inf\left\{k:~k\geq n_0~~\mbox{and}~~
{\mu}_{jk}\leq \frac{d_1^2k}{\tilde{a}_{j}^2} ~~\mbox{and}~~ \v {A_j} \leq \left(\frac{d_2}{a_{p}}\right)^2\right\},
\end{align}
where ${\mu}_{jk}=\lambda_{max}[k\vesub{L}{j}\vess{\Sigma}{jk}{-1}\vesub{L}{j}]$, $\vesub{\Sigma}{jk}=\sum_{i=1}^k \vesub{x}{ji}\{\dot\mu(\vesub{x}{ji}^{\top}\vesub{\beta}{j0})^2/\nu(\vesub{x}{ji}^{\top}\vesub{\beta}{j0})\}\vess{x}{ji}{\top}$, and $\lambda_{max}(A)$ denotes the maximum eigenvalue of matrix $A$.

In \eqref{Ndef-var}, $\mu_{jk}\leq (d_1^2k)/\tilde{a}_{j}^2$ controls the precision of the estimates for the parameters of interest, and $\v{Aj}\leq ({d_2}/{a_{p}})^2$ maintains the prediction accuracy of the model.
When $d_1$ and $d_2$ decrease, the sample size $\tilde{N}_j$ increases, and then we estimate  $\tvesub{\theta}{j\tilde{N}_j}$ with greater accuracy and prediction accuracy based on $A_j$. It follows from \eqref{Ndef-var} that practitioners can choose $d_1$ and $d_2$ based on their application needs.

Suppose we are at the $(k-1)$st stage, $k > n_0$,  of the sequential estimation procedure for the $j$th site, and have recruited $k-1$ samples into the estimation procedure.
If the inequalities in \eqref{Ndef-var} are satisfied with data set $C_{j{k-1}}$, then we stop recruiting new samples and report the current results.
Otherwise, we select an additional sample from the data pool of the $j$th site and use all $k$ samples in $C_{j{k}}$ to calculate new estimates $\tvesub{\beta}{jk}$ and $\nu_{jk}$. This  sequential recruiting process is repeated until the inequalities in \eqref{Ndef-var} are satisfied.
It has been proven that the estimates of parameters for GLM have uniform continuity in probability (u.c.i.p.) property \citep{Woodroofe1982,Chang-Martinsek,chang2011}. Moreover,  the property u.c.i.p. implies that the estimates are asymptotically normally distributed as the sample size goes to infinity. Thus, for each procedure $j$, $\tvesub{\theta}{j\tilde{N}_{j}}$ 
has the following asymptotic properties:
\begin{align}
&\sqrt{\tilde{N}_j}(\tvesub{\theta}{j}-\vesub{\theta}{0})\longrightarrow
N(0,{\vesub{L}{j}}\Sigma_{j}^{-1} {\vesub{L}{j}})
~~~~\text{\rm in distribution as }d_1\rightarrow 0,\nonumber\\
&\frac{\hat{A}_j-A_j}{\sqrt{v_{Aj}}}\longrightarrow
N(0,1)
~~~~\text{\rm in distribution as }d_2\rightarrow 0.\nonumber
\end{align}

As mentioned before, we independently conduct $M$ estimation procedures for constructing a confidence set of prefixed size for $\ve\theta$ in  (\ref{model1}) and
each procedure sequentially recruits samples from its corresponding data pool without replacement.
The fixed-sized (fixed-precision) feature of these $M$ estimates allows us to legitimately combine them into one final result with the desired statistical properties. We will state our integration method below.

\subsection{Combining sequential estimations from data centers}

Equation (\ref{Ndef-var}) states that  $\tilde{a}_{j}^2$ only depends on ${p}_0$ -- the number of 
the variables of interest.
Moreover, if  $\tilde{a}_{j}^2$  becomes larger, then we need to recruit more samples. 
Thus, we may adjust the sample proportion among $M$ sites   according to the data collection status for each site through the value of $\tilde{a}_{j}$ as long as
$\sum_{j=1}^M \tilde{a}_{j}^2 = a^2$.
When site $k$ has a much smaller number of subjects compared to others, we can take $\tilde{a}_{k}^2$ with a smaller value such that it can provide appropriate information with its limited data size via the predefined stopping criterion.

When all stopping rules $\tilde{N}_j$, $j=1,\cdots, M$, are fulfilled, we stop recruiting new data into our estimation procedures. 
Let 
\begin{eqnarray}
\hat{N} &=& \sum_{j=1}^M \tilde{N}_j,
\end{eqnarray}
and
\begin{eqnarray}
\hve{\theta} &=& \sum_{j=1}^M \rho_j\tvesub{\theta}{j\tilde{N}_j},
\end{eqnarray}
where $\rho_j=\tilde{N}_j/\hat{N}$.
Because these stopping times $\tilde{N}_j$, $j=1,\cdots,M$, are random, the sum of stopping time $\hat{N}$ indicates the total (random) samples used in $M$ processes together, and $\hve{\theta}$ is a ``randomly weighted'' estimate of $\vesub{\theta}{0}$ with random weights $\{\rho_j\}$, $j=1, \cdots, M$.

\begin{lemma}\label{Lemma-minvar}
Assume that for each $j$, $\{(\vesub{x}{ji}, y_{ji}), i \geq 1\}$ follows the GLM (\ref{meanfun}) and (\ref{varfun}), and Conditions {\em (A1)} and {\em (A2)} hold.
Then we have the estimate $\hve{\theta}$ with the random weights $\rho_j$ that achieves the minimal covariance asymptotically if all data centers have the same covariates and homogeneous covariance $\Sigma_1=\Sigma_2=\cdots=\Sigma_M$. 
\end{lemma}
The proof of Lemma 1 is in Supplementary. When all sites have homogeneous data with the same covariates,
the proposed random weight estimate achieves asymptotical efficiency under the weight combination of the parameter estimates from $M$ sites. Especially when $\tilde{a}_{j}$ are different, the proposed estimate has smaller asymptotical variance than the usual average of the parameter estimates from $M$ sites.

\begin{remark}
In practice, the data pool in each local site may contain some specific variables for their own, such as administrative purposes, which are usually not of interest for the analysis. Thus, focusing on a set of common variables among all sites is reasonable.
\end{remark}

Because $M$ procedures independently use their own data pools, and each estimate has the u.c.i.p. property, it follows that, as $d_1\rightarrow 0$,
\begin{align}\label{Tasynorm}
\sqrt{\hat{N}}\left(\sum_{j=1}^M\rho_j {\vesub{L}{j}}\Sigma_{j}^{-1} {\vesub{L}{j}}\right)^{-1/2}\left(\hve{\theta}-\vesub{\theta}{0}\right)\longrightarrow
N(0,\vesub{I}{p_0})
\hbox{ \rm in distribution}, 
\end{align}
where $\vesub{I}{p_0}$ is an identity matrix with rank $p_0$.
Equation (\ref{Tasynorm}) implies that
\begin{align}
(\hve{\theta}-\vesub{\theta}{0})^\top \tvesup{\Sigma}{-1}(\hve{\theta}-\vesub{\theta}{0})\longrightarrow \chi_{p_0}^2, \hbox{ as }d_1\rightarrow 0, \label{dist-var}
\end{align}
where $\tve{\Sigma}=\left(\sum_{j=1}^M\rho^2_j {\vesub{L}{j}}\Sigma_{j\tilde{N}_j}^{-1} {\vesub{L}{j}}\right)$. 
Let $\ve{Z}=(z_1,\cdots, z_{p_0})^{T}$, then 
\begin{align}\label{Rn-var}
R_{\hat{N}}\!\! =\!\left\{ \ve{Z} \in R^{p_0}\!\!:~~\!\!\frac{S_{\hat{N}}}{\hat{N}}\leq
\frac{d_1^2}{\mu_{\hat{N}}}\right\}
\end{align}
defines a confidence set for $\vesub{\theta}{0}$,
where $S_{\hat{N}}=(\ve{Z}-\hve{\theta})^{\top}\tvesup{\Sigma}{-1}(\ve{Z}-\hve{\theta})$,
$\mu_{\hat{N}}=\lambda_{max}\left(\hat{N}\tve{\Sigma}\right)$, and the length of the maximum axis of the confidence set $R_{\hat{N}}$ is $2d$.
For the stopping time $\hat{N}$ and confidence set $R_{\hat{N}}$, we have the following Theorem \ref {Th1}, and its proof is included in the Supplementary materials.
\begin{theorem}\label{Th1}
Assume that for each $j$, $\{(\vesub{x}{ji}, y_{ji}), i \geq 1\}$ follows the GLM
(\ref{meanfun}) and (\ref{varfun}), and Conditions 
{\em (A1)} and {\em (A2)} are satisfied. Then
\begin{align*}
& (i)\lim_{d_1\rightarrow 0}\frac{d_1^2\hat{N}}{a^2{\mu}}=1, \hbox{\rm almost surely},  \\
& (ii) \lim_{d_1\rightarrow 0} P(\vesub{\theta}{0} \in R_{\hat{N}})=1-\alpha, \\
& (iii) \lim_{d_1\rightarrow 0}\frac{d^2E(\hat{N})}{a^2{\mu}}=1, \nonumber
\end{align*}
where ${\mu}$ is the maximum eigenvalue of $\sum_{j=1}^M\rho_j {\vesub{L}{j}}\Sigma_{j}^{-1} {\vesub{L}{j}}$, and $a^2$ is the $1-\alpha$ quantile of $\chi_{p_0}^2$.
\end{theorem} 
 This theorem shows that  the proposed method has the prespecified coverage probability $1-\alpha$, and the ratio of the (random) total sample size to the best  ``theoretical''  (but unknown) size is asymptotically equal to 1. \citet{Chow1965} referred to these two properties  as {\it asymptotic consistency} and {\it asymptotic efficiency}, respectively.

The proposed method for aggregating the information among heteroscedastic data sites goes beyond the conventionally discussed challenges that drove the development of federated learning.   
In practice, the sizes and quality of data are different among different data sites. 
In a confidence set estimation scenario, it may not be possible to achieve the prefixed accuracy even when using all data in a site, which has only a small amount of data. With the proposed procedure, we can assign smaller $\tilde{a}_j$ to such sites and then effectively integrate all results from variant sites. Thus, the data usage is more efficient compared to that of other combining methods.

\begin{remark}
We can also modify the current methods by adopting a multiple-stage sequential method instead so that we can find more information about individual sites when there is little information available in advance. Please refer to \cite{park-chang2016} and the references therein. 
\end{remark}

\subsection{Adaptive sampling strategy}
Active learning is a long-standing topic in the machine learning literature. Researchers may use conventional experimental design criteria, such as some optimal design criteria, A-optima and D-optima criteria, for recruiting the most ``informative'' samples from the collected data {\citep{woods06, Deng2009, Montgomery2009, Zimu2020}}. 
 
Let $\mathbf A = \sum_{i=1}^k \vesub{x}{ji}\{\dot\mu(\vesub{x}{ji}^{\top}\vesub{\beta}{j0})^2/\nu(\vesub{x}{ji}^{\top}\vesub{\beta}{j0})\}\vess{x}{ji}{\top}$ for $j$th data site at the $k$th stage, where $\{\vesub{x}{ji}:i=1,...,k\}$ denotes the (active) set of the selected samples up to the $k$th stage. Let $\mathbf U$ be an inactive sample pool as in \cite{settles2010active} \citep[see also][]{Zimu2020, Liactive2020}.
The A-optima criterion will allocate a new observation $\vess{x}{j}{*}$ from the given data pool such that
 \begin{align*}
 \vess{x}{j}{*}={\mathrm{argmin}}_{x \in \mathbf{U}} \mathrm{tr}\{(\mathbf A +\ve{x} \{\dot\mu(\ve{x}^{\top}\vesub{\beta}{j0})^2/\nu(\ve{x}^{\top}\vesub{\beta}{j0})\}\vesup{x}{\top})^{-1}\},
 \end{align*}
where notation $\mathrm{tr}(\cdot)$ is the trace function.
Thus, we use the A-optima criterion to select a new observation at each stage in the modeling process and repeat this procedure until the stopping criterion (\ref{Ndef-var}) is satisfied. We then integrate results from all sites similar to those used in the random sampling procedure. Following the arguments of {\cite{wangchang13}}, we have the required asymptotic properties under an adaptive sampling scenario. The details are then omitted.

\begin{remark}
Although we have mainly discussed the confidence set estimation problems, this method can also accelerate the analysis of big data sets and not just for federated learning scenarios. It can be easily implemented on different computing frameworks due to its distributed computing feature. For example, we can simply partition a big data pool into M subdata sets and then fit the desired models separately using individual subdata sets. Thus, the proposed method goes beyond the original perspectives for developing federated learning from computing viewpoints.
\end{remark}

\section{Numerical studies}

\subsection{Simulation studies}

The performance of the proposed distributed sequential federated learning estimation (DSFL) method is investigated through simulation studies. We generate data using a logistic regression model:
\begin{eqnarray}\label{model}
P(Y=1|\ve X)=\mu(\vesup{X}{\top}\ve\beta)=\frac{\exp{(\beta_0+\vesup{X}{\top}\ve\beta)}}{1+\exp{(\vesup{X}{\top}\ve\beta)}},
\end{eqnarray}
where $\ve\beta \in R^{p_k-1}$ is the parameter vector of this model. 
We set $M=5$ for our simulation study,
and use real data to evaluate our methods with more sites later.
%

Below we list the parameters used in the simulation studies:
\begin{itemize}
\item[(1)] { Let {\it B1} denote the case that $\vesub{\beta}{k}=(\beta_{k0}, \vesup{\theta}{\top}, \vess{\beta}{k}{\top})^\top$, $\ve{\theta}=(2.0, 1.0)^\top$, $k=1,\cdots,5$ where $\ve{\theta}=(2.0, 1.0)^\top$ is the same for all sites, and {\it B2} denote the case that the parameter vectors are different among sites with $\beta_{k0}$, and $\vess{\beta}{k}{\top}$ as follows:}
\begin{eqnarray}
(\beta_{k0},\vesup{\theta}{\top},\vess{\beta}{1}{\top})^\top &=(-2,& 2, 1.0,  1.0, 0)^\top, \nonumber \\
 (\beta_{k0},\vesup{\theta}{\top},\vess{\beta}{2}{\top})^\top &=(-2,& 2, 1.0, 1.0, 0.5)^\top,\nonumber \\
 (\beta_{k0},\vesup{\theta}{\top},\vess{\beta}{3}{\top})^\top &=(-2,& 2, 1.0, 1.0, 0.5, 0)^\top,\nonumber\\ 
 (\beta_{k0},\vesup{\theta}{\top},\vess{\beta}{4}{\top})^\top &=(-1.5,& 2, 1.0, 1.0, 0)^\top,\nonumber\\
 (\beta_{k0},\vesup{\theta}{\top},\vess{\beta}{5}{\top})^\top &=(-2.5,& 2, 1.0, 1.0, 1.0)^\top.\nonumber  
\end{eqnarray}
Note that in addition to the values, the length of $\bm\beta_{3}$ is also different from the others. 
\item [(2)] For $p1$, $\rho_k=1/5$, for all $k=1, \cdots, 5$, and for $p2$, we have $\rho_1,\cdots,\rho_4 = 1/10$, and $\rho_5=6/10$.
\item[(3)] Random vector $X$ under $h1$ follows $N(0,{\rm {diag}}(\phi_{ki}, i=1,\cdots,p_{k}-1))$ with $p_k=5$, $\phi_{ki}=1$ for $i=1,\cdots,p_k-1, k=1,\cdots,5$. For $h2$, $X$ follows $N(0,{\rm {diag}}(\phi_{ki}, i=1,\cdots,p_{k}-1))$ with different $p_k$,  $\phi_{23}=\phi_{24}=4, \phi_{43}=\phi_{44}=2,\phi_{53}=\phi_{54}=4$, 
and the other $\phi_{ki}$ are equal to 1, for $k=1,\cdots,5$. 
\end{itemize}
In our studies with the synthesized data, 
the sizes of the confidence set for parameters $(\beta_1,\beta_2)$ are $d_1=0.4, 0.3$ and $0.2$, and the parameter for controlling the AUC of each model is $d_2=0.06, 0.05,  0.04$. We set the significance level at $\alpha = 0.05$, and the results were based on 200 replications for all cases.

Table \ref{tab1} presents the stopping times, coverage frequency (CF), and average AUC for models with either random (R) or adaptive (A) sample selection methods, using the parameter setup {\it{B1}}. 
The stopping time, $N$, increases as $d_1$ goes to 0, and then the coverage frequency approaches the theoretical value, 0.95. The procedures with adaptive sampling via the A-optimal design criterion have smaller stopping times than their random sampling counterparts. 
When $\rho=1/5$ for each site, as in $p1$, the stopping times for all 5 sites are similar. 
However, when the values of $\rho$ are different, as in $p2$, Sites 1-4 ($\rho_1=\cdots=\rho_4=0.1$) have much smaller sample sizes than Site 5 ($\rho_5=0.6$). This shows that we can control the size of the sample used at each site via choosing an appropriate $\rho$. 

Tables \ref{tab2} and \ref{tab2-1} show the absolute bias of the parameter estimation of interest from the proposed method, parameter estimates by combining 5 sites with equivalent weights, and each site for the R and A selection methods.
We found that compared with combining results from all 5 data sites, the results via single site data have much larger biases, and standard deviations.
When $d_1$ is small, the proposed method has similar biases and standard deviation under the $p1$ setup, compared to that of the method via straight average. However, under the $p2$ setup, the proposed method have smaller biases and standard deviations than the straight average method.   
The differences are even more obvious when an adaptive sampling strategy is adopted. As expected, the stopping times of all 5 sites are similar under $p1$, and vary under $p2$.

We use both $h1$ and $h2$ to study the heterogeneity effects due to other variables.
For each data site, regression parameters have different lengths with their values as described in $B2$ above.
We consider 4 combinations of parameters: $p1$ and $h1 ~(s1)$, $p2$ and $h1 ~(s2)$, $p1$ and $h2 ~(s3)$, and $p2$ and $h2 ~(s4)$. 
Table \ref{tab3} reports the stopping times, coverage frequency (CF), and average AUC of 5 sites with $d_1=0.2$ and $d_2=0.05$. Table \ref{tab4} shows the corresponding absolute bias for the estimates of the parameters of interest obtained via the proposed and straight average methods for each site. 
For cases $s2$ and $s4$, the proposed method has smaller biases and standard deviations than those of the simple average and methods using only single site data.

{
The results based on the synthesized data confirm that the proposed sequential federated learning yeilds a satisfactory result for estimating the parameters of $\beta_1$ and $\beta_2$ in all setups, and retains the demanding statistical properties in this study.}

\begin{table*}[h]
\caption{stopping times, AUC and coverage frequency with homogeneous covariates of each data site for the simulated data.}
\label{tab1}\tabcolsep=1pt\fontsize{6}{8}\selectfont
 \vskip 0pt
\par
\begin{center}
\begin{tabular}{ccccc cccc cccc}

&&&&\multicolumn{6}{c}{Stopping time}&&\\
\cline{5-10} 
&$d_2$&$d_1$&&$N$&$N_1$ &$N_2$& $N_3$& $N_4$ & $N_5$ && AUC &CF\\
R&0.06&0.4&p1&1127.420(116.651)&219.750(52.976)&227.640(58.507)&228.335(53.052)&227.215(54.243)&224.480(55.704)&&0.903(0.009)&0.915\\
&&&p2&1193.870(123.002)&141.160(31.880)&144.850(31.103)&148.820(33.206)&146.380(32.687)&612.660(98.263)&&0.907(0.009)&0.965\\
&&0.3&p1&1838.760(185.119)&361.895(83.340)&370.995(82.755)&370.225(78.068)&371.235(74.345)&364.410(86.074)&&0.901(0.008)&0.895\\
&&&p2&1862.495(173.527)&201.465(47.522)&198.705(48.032)&204.145(45.102)&207.550(50.823)&1050.630(140.413)&&0.903(0.008)&0.955\\
&&0.2&p1&3962.110(270.694)&795.065(113.044)&793.180(117.097)&781.050(118.549)&802.765(117.263)&790.050(121.204)&&0.902(0.005)&0.955\\
&&&p2&3954.730(259.459)&409.295(84.372)&407.690(79.709)&408.225(79.819)&410.225(90.593)&2319.295(208.466)&&0.901(0.007)&0.960\\
&0.05&0.4&p1&1180.350(110.364)&241.200(55.236)&231.700(43.742)&237.860(49.061)&234.910(51.000)&234.680(39.495)&&0.904(0.009)&0.920\\
&&&p2&1292.470(125.559)&170.245(42.899)&171.965(39.977)&173.190(40.141)&176.115(41.536)&600.955(107.197)&&0.908(0.008)&0.965\\
&&0.3&p1&1845.155(173.025)&377.325(84.753)&368.590(68.487)&369.100(78.389)&365.540(77.612)&364.600(79.501)&&0.902(0.008)&0.925\\
&&&p2&1918.810(150.884)&217.020(43.842)&214.135(41.127)&219.315(42.147)&218.955(42.511)&1049.385(139.777)&&0.905(0.007)&0.925\\
&&0.2&p1&3987.365(271.466)&807.735(121.610)&792.770(101.409)&796.120(131.255)&799.635(127.288)&791.105(126.285)&&0.902(0.005)&0.915\\
&&&p2&3950.750(252.432)&408.005(79.620)&399.740(77.757)&409.415(81.766)&400.175(77.379)&2333.415(211.235)&&0.901(0.006)&0.960\\
0.04&0.4&0.4&p1&1409.665(115.474)&283.315(46.822)&280.755(51.205)&285.885(54.894)&279.535(43.948)&280.175(50.207)&&0.907(0.008)&0.945\\
&&&p2&1592.525(158.412)&244.215(64.961)&256.545(72.169)&249.130(68.352)&247.605(62.649)&595.030(93.138)&&0.908(0.008)&0.955\\
&&0.3&p1&1899.810(140.624)&374.900(63.191)&378.050(61.751)&391.030(71.765)&378.685(67.319)&377.145(67.664)&&0.903(0.007)&0.920\\
&&&p2&2133.280(164.460)&268.430(52.070)&282.930(54.898)&275.985(51.456)&272.100(49.524)&1033.835(127.392)&&0.905(0.007)&0.960\\
&&0.2&p1&3953.785(267.599)&774.375(113.544)&786.030(122.170)&810.615(123.860)&798.300(118.161)&784.465(124.124)&&0.902(0.006)&0.935\\
&&&p2&3953.585(251.092)&411.455(69.285)&414.440(75.603)&410.125(70.166)&414.435(72.966)&2303.130(198.681)&&0.901(0.006)&0.940\\
A&0.06&0.4&p1&779.195(60.423)&157.370(27.523)&153.775(26.695)&153.010(24.152)&155.595(27.593)&159.445(30.977)&&0.897(0.006)&0.950\\
&&&p2&839.985(77.668)&117.980(21.386)&115.330(18.882)&116.465(21.355)&116.295(20.234)&373.915(59.929)&&0.901(0.006)&0.885\\
&&0.3&p1&1181.445(90.720)&238.465(43.166)&234.650(37.981)&230.065(40.849)&234.105(46.879)&244.160(40.479)&&0.893(0.006)&0.960\\
&&&p2&1231.420(97.393)&146.555(26.299)&146.260(25.736)&144.695(26.000)&147.710(24.105)&646.200(82.551)&&0.895(0.006)&0.920\\
&&0.2&p1&2458.665(151.224)&492.260(61.309)&498.565(62.355)&481.495(64.580)&488.935(63.845)&497.410(62.629)&&0.886(0.004)&0.965\\
&&&p2&2515.910(144.705)&259.490(47.387)&261.895(46.299)&261.305(47.419)&260.930(45.253)&1472.290(111.568)&&0.888(0.005)&0.950\\
&0.05&0.4&p1&872.290(58.272)&172.975(25.499)&172.095(25.739)&179.750(26.169)&176.220(27.657)&171.250(26.573)&&0.901(0.005)&0.930\\
&&&p2&972.265(83.208)&153.410(32.018)&145.595(30.334)&146.720(30.308)&148.950(33.610)&377.590(54.576)&&0.904(0.006)&0.945\\
&&0.3&p1&1203.885(83.544)&238.290(39.105)&237.205(34.790)&244.670(38.999)&240.440(41.670)&243.280(38.201)&&0.893(0.005)&0.925\\
&&&p2&1328.320(89.670)&166.100(24.505)&168.325(24.836)&170.980(28.018)&168.410(24.977)&654.505(72.764)&&0.898(0.005)&0.930\\
&&0.2&p1&2447.690(140.464)&489.660(63.909)&482.050(64.855)&491.925(65.949)&492.500(61.799)&491.555(64.902)&&0.885(0.004)&0.965\\
&&&p2&2551.055(149.139)&262.870(44.496)&269.955(44.021)&272.190(47.966)&271.170(40.905)&1474.870(115.197)&&0.890(0.005)&0.935\\
0.04&0.4&0.4&p1&1286.435(116.231)&259.770(52.710)&253.700(48.164)&257.625(52.416)&254.005(47.705)&261.335(52.356)&&0.900(0.006)&0.985\\
&&&p2&1379.530(125.618)&248.665(58.304)&255.315(63.360)&245.350(63.381)&249.430(58.984)&380.770(50.671)&&0.899(0.007)&0.995\\
&&0.3&p1&1420.685(84.562)&284.455(36.414)&284.250(39.525)&282.260(35.245)&284.255(35.636)&285.465(39.358)&&0.896(0.005)&0.955\\
&&&p2&1667.640(132.199)&253.015(53.753)&259.550(54.184)&257.760(49.870)&254.880(56.937)&642.435(78.943)&&0.896(0.006)&0.975\\
&&0.2&p1&2463.705(150.922)&495.770(68.153)&495.555(66.413)&488.955(64.063)&492.855(65.264)&490.570(65.247)&&0.886(0.004)&0.930\\
&&&p2&2657.900(139.943)&295.455(32.093)&300.855(38.706)&299.955(32.010)&301.330(32.319)&1460.305(115.925)&&0.891(0.004)&0.950\\

 \multicolumn{13}{l}{$^{*}$ Standard deviations are in parentheses. R and A stand for Random and Adaptive samplings, respectively.}
\end{tabular}
\end{center}
\end{table*}

\begin{table*}[!h]
\caption{Results of absolute bias of parameter estimate with random selection strategy and homogenuous covariates in each data center.}
\label{tab2}\tabcolsep=3pt\fontsize{8}{10}\selectfont
 \vskip 0pt
\par
\begin{center}
\begin{tabular}{cccc ccccc ccc}

&$d_2$&$d_1$&&&Proposed & Avearge & Site 1 &Site 2& Site 3 &Site 4& Site 5\\
&0.06&0.4&p1&$\beta_1$&0.141(0.104)&0.114(0.087)&0.250(0.180)&0.262(0.186)&0.253(0.179)&0.238(0.189)&0.257(0.174)\\
 && &&$\beta_2$&0.100(0.073)&0.089(0.067)&0.216(0.158)&0.200(0.130)&0.195(0.130)&0.196(0.134)&0.215(0.154)\\
&&&p2&$\beta_1$&0.129(0.100)&0.147(0.109)&0.291(0.231)&0.306(0.225)&0.305(0.222)&0.297(0.220)&0.145(0.112)\\
 && &&$\beta_2$&0.098(0.065)&0.110(0.079)&0.255(0.188)&0.256(0.177)&0.254(0.200)&0.251(0.193)&0.112(0.086)\\
&&0.3&p1&$\beta_1$&0.105(0.081)&0.092(0.073)&0.204(0.151)&0.214(0.157)&0.199(0.157)&0.181(0.147)&0.222(0.159)\\
 && &&$\beta_2$&0.076(0.056)&0.070(0.055)&0.156(0.112)&0.144(0.109)&0.141(0.101)&0.153(0.117)&0.175(0.132)\\
&&&p2&$\beta_1$&0.101(0.076)&0.109(0.084)&0.251(0.203)&0.249(0.186)&0.240(0.180)&0.261(0.200)&0.119(0.087)\\
 && &&$\beta_2$&0.072(0.054)&0.087(0.062)&0.208(0.148)&0.205(0.132)&0.205(0.159)&0.207(0.152)&0.088(0.066)\\
&&0.2&p1&$\beta_1$&0.063(0.047)&0.061(0.046)&0.132(0.101)&0.138(0.100)&0.138(0.105)&0.140(0.101)&0.140(0.112)\\
 && &&$\beta_2$&0.047(0.034)&0.046(0.035)&0.089(0.075)&0.096(0.074)&0.101(0.071)&0.100(0.078)&0.109(0.084)\\
&&&p2&$\beta_1$&0.061(0.045)&0.077(0.055)&0.197(0.151)&0.179(0.134)&0.174(0.134)&0.201(0.164)&0.078(0.060)\\
 && &&$\beta_2$&0.045(0.036)&0.060(0.046)&0.155(0.115)&0.145(0.106)&0.152(0.106)&0.151(0.115)&0.059(0.046)\\
&0.05&0.4&p1&$\beta_1$&0.136(0.103)&0.113(0.086)&0.252(0.196)&0.218(0.171)&0.227(0.188)&0.236(0.193)&0.223(0.177)\\
 && &&$\beta_2$&0.097(0.070)&0.092(0.066)&0.199(0.148)&0.176(0.126)&0.181(0.143)&0.192(0.146)&0.167(0.125)\\
&&&p2&$\beta_1$&0.124(0.089)&0.144(0.098)&0.280(0.222)&0.293(0.221)&0.304(0.223)&0.277(0.218)&0.154(0.126)\\
 && &&$\beta_2$&0.084(0.063)&0.096(0.073)&0.242(0.183)&0.236(0.186)&0.220(0.149)&0.228(0.169)&0.112(0.087)\\
&&0.3&p1&$\beta_1$&0.101(0.071)&0.088(0.066)&0.212(0.152)&0.188(0.139)&0.205(0.151)&0.201(0.150)&0.198(0.155)\\
 && &&$\beta_2$&0.074(0.054)&0.070(0.052)&0.156(0.116)&0.135(0.108)&0.146(0.107)&0.152(0.120)&0.145(0.107)\\
&&&p2&$\beta_1$&0.093(0.071)&0.100(0.074)&0.244(0.174)&0.220(0.162)&0.249(0.175)&0.229(0.165)&0.119(0.094)\\
 && &&$\beta_2$&0.068(0.051)&0.080(0.057)&0.204(0.152)&0.199(0.150)&0.198(0.126)&0.188(0.137)&0.087(0.065)\\
&&0.2&p1&$\beta_1$&0.067(0.053)&0.061(0.049)&0.142(0.107)&0.117(0.091)&0.142(0.119)&0.143(0.104)&0.143(0.108)\\
 && &&$\beta_2$&0.046(0.035)&0.044(0.034)&0.105(0.074)&0.092(0.076)&0.094(0.080)&0.110(0.074)&0.102(0.079)\\
&&&p2&$\beta_1$&0.057(0.050)&0.073(0.052)&0.191(0.137)&0.174(0.136)&0.184(0.137)&0.171(0.138)&0.077(0.063)\\
 && &&$\beta_2$&0.045(0.032)&0.056(0.044)&0.131(0.100)&0.145(0.104)&0.159(0.110)&0.130(0.109)&0.059(0.043)\\
&0.04&0.4&p1&$\beta_1$&0.137(0.097)&0.128(0.088)&0.220(0.176)&0.224(0.171)&0.234(0.198)&0.225(0.183)&0.212(0.174)\\
 && &&$\beta_2$&0.090(0.069)&0.085(0.067)&0.175(0.137)&0.185(0.146)&0.179(0.134)&0.176(0.138)&0.197(0.140)\\
&&&p2&$\beta_1$&0.115(0.090)&0.150(0.115)&0.270(0.239)&0.301(0.233)&0.292(0.249)&0.306(0.227)&0.144(0.101)\\
 && &&$\beta_2$&0.083(0.065)&0.099(0.079)&0.202(0.156)&0.208(0.178)&0.205(0.176)&0.212(0.194)&0.108(0.083)\\
&&0.3&p1&$\beta_1$&0.109(0.073)&0.093(0.064)&0.180(0.127)&0.186(0.136)&0.200(0.150)&0.178(0.141)&0.185(0.142)\\
 && &&$\beta_2$&0.075(0.053)&0.069(0.050)&0.143(0.113)&0.155(0.119)&0.147(0.114)&0.158(0.117)&0.170(0.118)\\
&&&p2&$\beta_1$&0.092(0.073)&0.110(0.088)&0.216(0.176)&0.247(0.182)&0.238(0.190)&0.249(0.172)&0.118(0.080)\\
 && &&$\beta_2$&0.068(0.052)&0.077(0.062)&0.174(0.139)&0.180(0.147)&0.180(0.148)&0.172(0.144)&0.086(0.064)\\
&&0.2&p1&$\beta_1$&0.064(0.048)&0.061(0.046)&0.136(0.093)&0.140(0.112)&0.146(0.107)&0.134(0.098)&0.145(0.103)\\
 && &&$\beta_2$&0.050(0.034)&0.049(0.034)&0.109(0.079)&0.103(0.076)&0.105(0.077)&0.107(0.076)&0.110(0.085)\\
&&&p2&$\beta_1$&0.061(0.046)&0.077(0.056)&0.160(0.118)&0.187(0.136)&0.182(0.135)&0.172(0.128)&0.076(0.062)\\
 && &&$\beta_2$&0.051(0.036)&0.060(0.045)&0.132(0.093)&0.137(0.112)&0.138(0.115)&0.138(0.104)&0.063(0.044)\\

 \multicolumn{11}{l}{$^{*}$ Standard deviations are in parentheses.}
\end{tabular}
\end{center}
\end{table*}

\begin{table*}[!h]
\caption{Results of absolute bias of parameter estimate with adaptive selection strategy and homogenuous covariates in each data center.}
\label{tab2-1}\tabcolsep=3pt\fontsize{8}{10}\selectfont
 \vskip 0pt
\par
\begin{center}
\begin{tabular}{cccc ccccc ccc}

&$d_2$&$d_1$&&&Proposed & Avearge & Site 1 &Site 2& Site 3 &Site 4& Site 5\\
&0.06&0.4&p1&$\beta_1$&0.140(0.098)&0.114(0.086)&0.248(0.177)&0.232(0.170)&0.225(0.165)&0.238(0.172)&0.264(0.191)\\
 && &&$\beta_2$&0.075(0.057)&0.067(0.051)&0.167(0.113)&0.139(0.107)&0.149(0.114)&0.139(0.101)&0.161(0.119)\\
&&&p2&$\beta_1$&0.149(0.114)&0.182(0.127)&0.341(0.248)&0.308(0.242)&0.322(0.244)&0.346(0.263)&0.165(0.120)\\
 && &&$\beta_2$&0.097(0.071)&0.113(0.082)&0.220(0.164)&0.212(0.163)&0.209(0.167)&0.211(0.159)&0.109(0.081)\\
&&0.3&p1&$\beta_1$&0.097(0.072)&0.089(0.062)&0.190(0.148)&0.179(0.129)&0.195(0.148)&0.208(0.161)&0.192(0.141)\\
 && &&$\beta_2$&0.053(0.043)&0.049(0.040)&0.124(0.091)&0.114(0.084)&0.118(0.089)&0.119(0.090)&0.122(0.091)\\
&&&p2&$\beta_1$&0.107(0.070)&0.122(0.085)&0.261(0.196)&0.258(0.188)&0.247(0.184)&0.259(0.191)&0.122(0.100)\\
 && &&$\beta_2$&0.065(0.050)&0.079(0.056)&0.174(0.139)&0.161(0.127)&0.169(0.130)&0.160(0.123)&0.078(0.062)\\
&&0.2&p1&$\beta_1$&0.065(0.049)&0.064(0.046)&0.126(0.091)&0.133(0.100)&0.136(0.108)&0.133(0.104)&0.129(0.097)\\
 && &&$\beta_2$&0.039(0.028)&0.037(0.029)&0.080(0.057)&0.083(0.061)&0.084(0.067)&0.084(0.060)&0.085(0.068)\\
&&&p2&$\beta_1$&0.061(0.043)&0.077(0.057)&0.197(0.141)&0.178(0.138)&0.193(0.137)&0.188(0.145)&0.075(0.052)\\
 && &&$\beta_2$&0.045(0.032)&0.055(0.038)&0.128(0.097)&0.124(0.098)&0.120(0.087)&0.116(0.086)&0.056(0.038)\\
&0.05&0.4&p1&$\beta_1$&0.127(0.099)&0.118(0.090)&0.220(0.168)&0.218(0.159)&0.266(0.182)&0.239(0.193)&0.202(0.149)\\
 && &&$\beta_2$&0.078(0.060)&0.075(0.056)&0.139(0.107)&0.144(0.118)&0.171(0.119)&0.152(0.119)&0.142(0.110)\\
&&&p2&$\beta_1$&0.144(0.109)&0.186(0.129)&0.308(0.232)&0.327(0.258)&0.349(0.255)&0.342(0.292)&0.155(0.114)\\
 && &&$\beta_2$&0.075(0.057)&0.091(0.070)&0.193(0.153)&0.174(0.145)&0.184(0.148)&0.190(0.149)&0.094(0.075)\\
&&0.3&p1&$\beta_1$&0.099(0.070)&0.085(0.064)&0.190(0.136)&0.178(0.124)&0.197(0.141)&0.203(0.148)&0.184(0.149)\\
 && &&$\beta_2$&0.062(0.044)&0.057(0.040)&0.125(0.093)&0.121(0.090)&0.128(0.096)&0.126(0.094)&0.123(0.096)\\
&&&p2&$\beta_1$&0.102(0.076)&0.127(0.088)&0.239(0.173)&0.262(0.192)&0.276(0.221)&0.252(0.182)&0.114(0.088)\\
 && &&$\beta_2$&0.062(0.045)&0.073(0.058)&0.154(0.118)&0.157(0.115)&0.161(0.138)&0.157(0.117)&0.073(0.051)\\
&&0.2&p1&$\beta_1$&0.060(0.043)&0.058(0.044)&0.129(0.101)&0.136(0.102)&0.136(0.105)&0.125(0.094)&0.132(0.103)\\
 && &&$\beta_2$&0.038(0.028)&0.038(0.028)&0.085(0.071)&0.087(0.066)&0.087(0.066)&0.085(0.059)&0.087(0.069)\\
&&&p2&$\beta_1$&0.070(0.050)&0.084(0.064)&0.199(0.136)&0.193(0.135)&0.205(0.140)&0.171(0.135)&0.080(0.055)\\
 && &&$\beta_2$&0.044(0.029)&0.053(0.040)&0.125(0.092)&0.110(0.083)&0.126(0.095)&0.115(0.087)&0.051(0.039)\\
&0.04&0.4&p1&$\beta_1$&0.111(0.087)&0.122(0.093)&0.213(0.180)&0.217(0.187)&0.229(0.176)&0.236(0.181)&0.201(0.172)\\
 && &&$\beta_2$&0.065(0.053)&0.070(0.055)&0.134(0.116)&0.139(0.108)&0.123(0.104)&0.134(0.109)&0.123(0.101)\\
&&&p2&$\beta_1$&0.105(0.077)&0.134(0.094)&0.237(0.203)&0.237(0.231)&0.274(0.241)&0.234(0.219)&0.159(0.119)\\
 && &&$\beta_2$&0.063(0.048)&0.080(0.059)&0.149(0.125)&0.134(0.129)&0.159(0.156)&0.159(0.142)&0.092(0.065)\\
&&0.3&p1&$\beta_1$&0.096(0.073)&0.091(0.068)&0.190(0.135)&0.186(0.142)&0.168(0.133)&0.186(0.128)&0.189(0.145)\\
 && &&$\beta_2$&0.057(0.044)&0.055(0.041)&0.111(0.092)&0.112(0.090)&0.105(0.085)&0.117(0.086)&0.118(0.088)\\
&&&p2&$\beta_1$&0.092(0.068)&0.115(0.082)&0.198(0.164)&0.222(0.197)&0.218(0.171)&0.226(0.177)&0.120(0.099)\\
 && &&$\beta_2$&0.056(0.043)&0.070(0.052)&0.138(0.107)&0.134(0.114)&0.134(0.116)&0.150(0.115)&0.080(0.056)\\
&&0.2&p1&$\beta_1$&0.066(0.052)&0.062(0.051)&0.143(0.108)&0.144(0.105)&0.134(0.103)&0.145(0.098)&0.133(0.105)\\
 && &&$\beta_2$&0.041(0.031)&0.040(0.029)&0.085(0.065)&0.092(0.068)&0.083(0.063)&0.085(0.061)&0.089(0.066)\\
&&&p2&$\beta_1$&0.061(0.044)&0.074(0.057)&0.158(0.122)&0.178(0.136)&0.177(0.137)&0.170(0.120)&0.072(0.057)\\
 && &&$\beta_2$&0.042(0.031)&0.053(0.035)&0.113(0.083)&0.114(0.083)&0.121(0.086)&0.112(0.085)&0.057(0.043)\\

 \multicolumn{11}{l}{$^{*}$ Standard deviations are in parentheses.}
\end{tabular}
\end{center}
\end{table*}

\begin{table*}[h]
\caption{Simulation results about stopping times, AUC and coverage frequency with heterogeneity covariates, $d_1=0.2$ and $d_2=0.05$.}
\label{tab3}\tabcolsep=2pt\fontsize{7}{9}\selectfont
 \vskip 0pt
\par
\begin{center}
\begin{tabular}{ccccc cccc cccc}

&&&&\multicolumn{6}{c}{Stopping time}&&\\
\cline{5-10} 
&&&&$N$&$N_1$ &$N_2$& $N_3$& $N_4$ & $N_5$ && AUC &CF\\
R&&&s1&3973.980(276.329)&792.765(118.450)&798.200(117.615)&781.180(113.771)&717.270(112.102)&884.565(141.991)&&0.902(0.006)&0.950\\
&&&s2&4218.450(282.242)&420.235(83.695)&413.230(82.100)&409.250(82.803)&370.620(69.013)&2605.115(250.404)&&0.903(0.007)&0.955\\
&&&s3&4215.865(261.601)&796.765(110.019)&867.045(141.630)&799.680(120.935)&771.295(103.102)&981.080(139.137)&&0.916(0.005)&0.940\\
&&&s4&4579.100(284.174)&405.350(86.785)&453.955(87.486)&409.570(82.998)&392.575(74.388)&2917.650(231.654)&&0.916(0.006)&0.965\\
A&&&s1&2479.915(132.460)&503.405(63.221)&488.560(67.064)&509.740(59.487)&483.295(57.845)&494.915(64.279)&&0.885(0.004)&0.960\\
&&&s2&2574.330(160.415)&267.140(46.490)&263.540(44.746)&274.190(45.693)&262.870(41.629)&1506.590(121.938)&&0.888(0.005)&0.905\\
&&&s3&2446.545(149.305)&494.795(64.300)&476.625(63.512)&500.510(64.792)&482.090(58.726)&492.525(61.964)&&0.889(0.004)&0.935\\
&&&s4&2513.765(149.527)&268.940(48.173)&260.395(44.636)&267.365(42.222)&251.965(45.746)&1465.100(113.487)&&0.893(0.005)&0.940\\

 \multicolumn{12}{l}{$^{*}$ Standard deviations are in parentheses. R and A stand for Random and Adaptive samplings, respectively.}
\end{tabular}
\end{center}
\end{table*}

\begin{table*}[!h]
\caption{Results of absolute bias of parameter estimate with with heterogeneity covariates, $d_1=0.2$ and $d_2=0.05$.}
\label{tab4}\tabcolsep=3pt\fontsize{9}{11}\selectfont
 \vskip 0pt
\par
\begin{center}
\begin{tabular}{ccccc cccc ccc}

&&&&&Proposed & Avearge & Site 1 &Site 2& Site 3 &Site 4& Site 5\\

R&&&s1&$\beta_1$&0.065(0.047)&0.061(0.047)&0.137(0.104)&0.138(0.098)&0.127(0.097)&0.138(0.103)&0.146(0.108)\\
& & &&$\beta_2$&0.048(0.037)&0.047(0.037)&0.099(0.069)&0.106(0.083)&0.104(0.082)&0.105(0.077)&0.111(0.081)\\
&&&s2&$\beta_1$&0.061(0.051)&0.073(0.054)&0.195(0.136)&0.191(0.142)&0.196(0.138)&0.174(0.133)&0.082(0.063)\\
 && &&$\beta_2$&0.047(0.034)&0.057(0.041)&0.148(0.113)&0.140(0.103)&0.131(0.097)&0.153(0.121)&0.067(0.049)\\
&&&s3&$\beta_1$&0.062(0.048)&0.058(0.043)&0.133(0.102)&0.145(0.112)&0.138(0.095)&0.139(0.098)&0.128(0.095)\\
 && &&$\beta_2$&0.046(0.034)&0.045(0.033)&0.096(0.073)&0.100(0.082)&0.099(0.078)&0.095(0.073)&0.108(0.076)\\
&&&s4&$\beta_1$&0.058(0.046)&0.069(0.053)&0.204(0.137)&0.180(0.132)&0.177(0.150)&0.179(0.131)&0.073(0.053)\\
 && &&$\beta_2$&0.045(0.033)&0.058(0.043)&0.162(0.126)&0.146(0.103)&0.149(0.107)&0.154(0.108)&0.054(0.039)\\
A&&&s1&$\beta_1$&0.058(0.043)&0.054(0.042)&0.127(0.105)&0.135(0.112)&0.125(0.089)&0.125(0.091)&0.131(0.109)\\
& & &&$\beta_2$&0.040(0.029)&0.039(0.029)&0.087(0.062)&0.085(0.069)&0.081(0.061)&0.078(0.058)&0.081(0.062)\\
&&&s2&$\beta_1$&0.067(0.054)&0.082(0.065)&0.198(0.134)&0.189(0.136)&0.182(0.146)&0.181(0.133)&0.078(0.059)\\
 && &&$\beta_2$&0.045(0.034)&0.054(0.040)&0.119(0.091)&0.122(0.089)&0.124(0.091)&0.115(0.087)&0.058(0.041)\\
&&&s3&$\beta_1$&0.062(0.051)&0.060(0.049)&0.133(0.102)&0.139(0.098)&0.132(0.107)&0.122(0.092)&0.129(0.096)\\
 && &&$\beta_2$&0.036(0.031)&0.036(0.030)&0.082(0.068)&0.072(0.063)&0.083(0.062)&0.084(0.059)&0.077(0.061)\\
&&&s4&$\beta_1$&0.065(0.049)&0.080(0.057)&0.216(0.145)&0.191(0.132)&0.175(0.126)&0.193(0.138)&0.074(0.057)\\
 && &&$\beta_2$&0.041(0.032)&0.051(0.035)&0.117(0.093)&0.116(0.091)&0.116(0.088)&0.117(0.091)&0.051(0.041)\\

 \multicolumn{12}{l}{$^{*}$ Standard deviations are in parentheses. R and A stand for Random and Adaptive samplings, respectively.}
\end{tabular}
\end{center}
\end{table*}

\subsection{COVID-19 data}

We use a COVID-19 data set collected in Mexico by Mexican health authorities to demonstrate our method. Please note that this data set may be updated all any time. We downloaded this dataset in April 2021, and it contained 6659184 registers of the COVID-19 patients at that time. The data were obtained from an epidemiological study of  suspected cases of the viral respiratory disease at that time. 
The data were identified in the medical units of the 32 health sectors.Outpatients and
inpatients were identified based on the clinical diagnosis at admission. The data set includes some personal and health information of the study subjects, such as gender, 
age, and medical history,
including pneumonia, diabetes, chronic obstructive pulmonary(COPD), asthma, immunosuppression (Immu), hypertension, cardiovascular disease (Card), 
chronic renal failure (CRF), other diseases diagnoses, as well as other factors such as obesity
smoking status, exposure to other cases of SARS CoV-2(EOC) and COVID-19 status (positive or negative). We treat COVID-19 status as the response variable and the others as covariates in our analysis. 
Other variables, such as ID, birthplace of subject, sector catalog, type of care, and date of onset of symptoms, are not included in our analysis.
All variables in this dataset except age, are binary, where 1 represents ``Y", and 0 represents ``N", and for gender, 1 indicates females, and 0 indicates males.

In this numerical study, we aim to find the association between COVID-19 status (positive or negative) and the other variables. In addition, some variables are of particular interest. For example, one research question of interest was as follows: are subjects with diabetes or obesity more likely to be affected with COVID-19  compared with other subjects?
To assess this, we adopt a logistic regression model for our analysis. 

After omitting subjects with missing values, 5816861 subjects remains in the dataset, and the range of sample sizes of 32 different sites are from 21746 to 2396133.  Because of the huge differences in sample size among the different sites, we use 2 sampling strategies: (C1) equal proportion among all sites; that is, each site contributes 1/32, of the samples, and (C2) 1/100 for sites 4, 6, 7, 18, and equal proportion, 6/175, for the rest. The reason we use strategy (C2) is that the total data sizes of sites 4, 6, 7, and 18 are less than 30000. By comparing the results based on these two sampling strategies, we can assess the strength of the proposed method.
 
When maintaining the confidentiality of personal health information and minimizing the data communication burden are essential, each health sector uses its own data to build a logistics model.  
As mentioned before, we separately train a logistic model for each health sector, report the stopping time and parameter estimates, and use the AUC to control the prediction performance of each model. We then integrate all the results as described previously.

For comparison purposes (the baseline model), we simply use all data to estimate parameters and then use the estimates and AUC value as references.
We consider three parameter combinations to illustrate our method as follows:
\begin{itemize}
    \item[]All:  all parameter variables are equally important;
    \item []Part 1: Only five variables,  pneumonia, COPD, asthma, CRF and EOC, are of interest;
    \item[]Part 2: The 10-variable case. Suppose that variables gender, age, diabetes, asthma, hypertension, other diseases diagnoses, cardiovascular, obesity, CRF and smoking status are of interest.
\end{itemize}

Tables \ref{tabr3} and \ref{tabr4} report the stopping times and AUC obtained by applying the proposed method to the COVID-19 data.
Table \ref{tabr3} shows that the proposed method yields larger AUCs than the baseline model. 
The models with 3 different variable combinations have similar AUCs because we set $d_2=0.05$ for the AUC for the 3 different variable sets.
Models with a random selection strategy usually require more than 100000 samples before the stopping criteria are fulfilled. Thus, sites with smaller data sizes, such as sites 4, 6, 7 and 18, could not achieve the preset precision of parameter estimation with their own data.  
Models using an adaptive selection method usually reach the stopping criteria with much smaller sample sizes. 

When the number of variables of interest becomes large, the stopping time also increases. Hence, if all parameters are of interest, then the stopping times, the required sample sizes, are much larger than those of Part 1 and Part 2. 
From the definition of the stopping criterion, it is clear that when $d_1$ is small, the proposed sequential methods would demand more samples. Table \ref{tabr3} shows that designs C1 (equal proportions) and C2 (different proportions) have similar total stopping times. However, Table \ref{tabr4} shows that under C2, Sites 4, 6, 7 and 18 use much smaller sample sizes than under C1, because C2 provides a smaller proportion for these sites. One of the advantages of early stopping for these cases is better control the samples used for modeling even for sites with limited samples; especially when adopting an adaptive sampling method, we can even ``guarantee'' the quality of samples.

Table \ref{tabr1} reports parameter estimates for COVID-19 data with equal proportions C1. 
We only list estimates of parameters of interest; hence, for Part 1, only estimates for the variables pneumonia, COPD, asthma, CRF, and EOC are reported.
These tables also show that the estimates of the models  focusing only on the parameters of interest are consistent with those of the baseline model. 

Table \ref{tabr2} shows the results for the different proportions C2, which lead to a similar conclusion with the equal proportion C1. Hence, we do not go into the details here. Based on Tables \ref{tabr1} and \ref{tabr2},  we find that adaptive selection yields parameter estimates consistent with those of the models using random selection. However, Table \ref{tabr3} shows that the models with adaptive sampling use much smaller sample sizes. All the numerical results based on this COVID-19 data analysis confirm that the proposed distributed sequential federated learning method performs well, and has good potential to be extended for other applications.

We briefly summarize the results of our analysis using the COVID-19 data below. However, please note that this was not a well-designed study, and we mainly use it to illustrate our method. 
Comparing the results of Part 2 with that of the baseline model, we found that the estimates of age, diabetes and obesity in both models are significantly greater than 0, which may indicate that elderly subjects with diabetes or obesity are more likely to be infected. In addition, based on this data analysis, we find that the parameter estimates for gender, Card and smoking status are significantly less than 0, which may suggest that female subjects with cardiovascular disease and a history of smoking have smaller probability of being infected. 
The results of our analysis also shows that the parameter estimates for pneumonia and EOC are statistically significantly greater than 0, and those for asthma and CRF are significantly less than 0.  These results suggest that subjects with pneumonia and EOC and without asthma and CRF are more likely to be infected.  Similar results for age, diabetes, obesity, gender, smoking status and asthma on COVID-19 inflection were also found in the studies of \citet{Hernandez2020,Rashedi2020,Louis2020,Liu2020,Memon2022}.

\begin{remark}
    The proposed method determines ``how many data" and ``which subdata-sets" selected from  heterogeneous sources of data when a stopping criterion is satisfied. During sequential procedures, in addition to requiring that the estimation of parameters of interest achieves the prespecified level of accuracy, the method guarantees that the model achieves the prespecified precision, such as prediction for the classification model.
  
\end{remark}

\begin{remark}
    We construct individual sequential sampling strategies considering varying data sizes of different data centers, to avoid sequential procedures with small data sizes that can not be stopped even when all data run out.
\end{remark}

\begin{table*}[t]
\caption{Stopping times and AUC for Covid data with $d_2=0.05$.}
\label{tabr3}
\tabcolsep=3pt\fontsize{9}{11}\selectfont
 \vskip 0pt
\par
\begin{center}
\begin{tabular}{cccccc cccc cc}

&&&\multicolumn{3}{c}{Stopping time}&&\multicolumn{3}{c}{AUC}&&\\
\cline{4-6} \cline{8-10}
&&$d_1$&All parameters& Part 1 & Part 2 & & All parameters& Part 1 & Part 2 && Baseline\\
R&C1&0.4&121680&51780&55580&&0.627&0.633&0.631&&0.598\\
&&0.3&199380&84280&91580&&0.625&0.629&0.627&&0.598\\
&&0.2&432780&170680&183380&&0.622&0.625&0.625&&0.598\\
&C2&0.4&123580&51580&56380&&0.628&0.635&0.632&&0.598\\
&&0.3&203280&84580&92680&&0.626&0.632&0.629&&0.598\\
&&0.2&434780&172880&185380&&0.622&0.626&0.627&&0.598\\
A&C1&0.4&16940&16250&16390&&0.673&0.673&0.672&&0.598\\
&&0.3&18610&16480&16990&&0.668&0.672&0.672&&0.598\\
&&0.2&27270&18020&20480&&0.662&0.670&0.666&&0.598\\
&C2&0.4&17080&16280&16450&&0.673&0.672&0.672&&0.598\\
&&0.3&18750&16550&17100&&0.670&0.672&0.672&&0.598\\
&&0.2&26920&18230&20720&&0.663&0.670&0.668&&0.598\\

 \multicolumn{11}{l}{$^{*}$ R and A stand for Random and Adaptive samplings, respectively.}
\end{tabular}
\end{center}
\end{table*}

\begin{table*}[ht]
\caption{Stopping times of centers 4, 6, 7, and 18 with data size less than 30000 for Covid data.}
\label{tabr4}\tabcolsep=3pt\fontsize{9}{11}\selectfont
 \vskip 0pt
\par
\begin{center}
\begin{tabular}{ccccc ccccc cccc ccccc}

&&&\multicolumn{4}{c}{All parameters}&&\multicolumn{4}{c}{Part 1}&&\multicolumn{4}{c}{Part 2}\\
\cline{4-7} \cline{9-12} \cline{14-17}
&&$d_1$&Site 4 & Site 6 & Site 7 & Site 18 & & Site 4 & Site 6 & Site 7 & Site 18 & &Site 4 & Site 6 & Site 7 & Site 18\\
R&C1&0.4&3915&2515&4415&1615&&2015&915&2315&515&&2515&1115&1515&715\\
&&0.3&5815&4615&6715&3315&&3715&1615&3815&1015&&3815&1815&2615&1215\\
&&0.2&14715&12215&12215&7615&&5215&3215&6815&2215&&8415&3915&5715&2715\\
&C2&0.4&1315&1115&1715&715&&615&515&715&515&&1315&615&715&515\\
&&0.3&2515&1815&2615&1015&&915&615&1115&515&&1915&815&915&515\\
&&0.2&4215&3615&5115&2015&&2515&1215&2615&715&&2915&1415&1915&1015\\
A &C1&0.4&525&445&385&465&&515&445&385&465&&515&445&385&465\\
&&0.3&605&455&555&465&&515&445&385&465&&545&445&385&465\\
&&0.2&1025&805&1025&705&&585&445&455&465&&735&505&595&495\\
&C2&0.4&515&445&385&465&&515&445&385&465&&515&445&385&465\\
&&0.3&515&445&385&465&&515&445&385&465&&515&445&385&465\\
&&0.2&545&445&425&465&&515&445&385&465&&515&445&385&465\\

 \multicolumn{11}{l}{$^{*}$ R and A stand for Random and Adaptive samplings, respectively.}
\end{tabular}
\end{center}
\end{table*}

\begin{table*}[ht]
\caption{Parameter estimation for CoVid data with $d_2=0.05$ and equal proportion C1.}
\label{tabr1}\tabcolsep=1pt\fontsize{6}{9}\selectfont
 \vskip 0pt
\par
\begin{center}
\begin{tabular}{ccccc cccc cccc cc ccc}

&&$d_1$&&Gender&  Pneumonia & Age & Diabetes& COPD & Asthma & Immu & Hypertension & Other disease & Card & Obesity & CRF & Smoke & EOC\\

R&All&0.4& Est.&-0.162&1.085&0.013&0.155&-0.252&-0.160&-0.374&0.026&-0.120&-0.337&0.234&-0.301&-0.095&0.275\\
&&&Sd&0.013&0.027&0.000&0.023&0.080&0.043&0.083&0.021&0.055&0.058&0.019&0.060&0.028&0.013\\
&&0.3& Est.&-0.169&1.097&0.012&0.159&-0.294&-0.150&-0.295&0.033&-0.124&-0.312&0.242&-0.275&-0.131&0.267\\
&&&Sd&0.010&0.021&0.000&0.018&0.060&0.034&0.062&0.016&0.041&0.044&0.015&0.046&0.021&0.010\\
&&0.2& Est.&-0.163&1.092&0.012&0.148&-0.276&-0.130&-0.331&0.032&-0.138&-0.265&0.236&-0.356&-0.182&0.256\\
&&&Sd&0.007&0.014&0.000&0.012&0.039&0.022&0.042&0.011&0.028&0.030&0.010&0.031&0.015&0.007\\
&Part1 &0.4& Est.&-&1.089&-&-&-0.060&-0.158&-&-&-&-&-&-0.375&-&0.279\\
&&&Sd&-&0.042&-&-&0.129&0.067&-&-&-&-&-&0.097&-&0.020\\
&&0.3& Est.&-&1.122&-&-&-0.231&-0.200&-&-&-&-&-&-0.296&-&0.297\\
&&&Sd&-&0.033&-&-&0.103&0.052&-&-&-&-&-&0.074&-&0.016\\
&&0.2& Est.&-&1.108&-&-&-0.305&-0.139&-&-&-&-&-&-0.248&-&0.282\\
&&&Sd&-&0.023&-&-&0.067&0.036&-&-&-&-&-&0.050&-&0.011\\
&Part 2&0.4& Est.&-0.165&-&0.013&0.160&-&-&-&0.026&-0.180&-0.346&0.209&-0.308&-0.116&-\\
&&&Sd&0.019&-&0.001&0.034&-&-&-&0.030&0.081&0.091&0.028&0.089&0.041&-\\
&&0.3& Est.&-0.172&-&0.012&0.182&-&-&-&0.027&-0.151&-0.334&0.220&-0.317&-0.121&-\\
&&&Sd&0.014&-&0.001&0.026&-&-&-&0.024&0.063&0.068&0.022&0.069&0.032&-\\
&&0.2& Est.&-0.166&-&0.013&0.161&-&-&-&0.019&-0.134&-0.294&0.224&-0.298&-0.126&-\\
&&&Sd&0.010&-&0.000&0.019&-&-&-&0.017&0.044&0.046&0.015&0.048&0.022&-\\
A&All&0.4& Est.&-0.174&1.100&0.014&0.099&-0.236&-0.082&-0.427&0.037&-0.267&-0.391&0.235&-0.280&-0.170&0.441\\
&&&Sd&0.034&0.051&0.001&0.048&0.065&0.062&0.066&0.045&0.062&0.063&0.044&0.063&0.053&0.036\\
&&0.3& Est.&-0.186&1.034&0.014&0.098&-0.265&-0.102&-0.350&0.036&-0.258&-0.357&0.265&-0.231&-0.191&0.451\\
&&&Sd&0.032&0.045&0.001&0.043&0.054&0.052&0.055&0.041&0.052&0.053&0.040&0.053&0.047&0.034\\
&&0.2& Est.&-0.173&0.915&0.014&0.142&-0.215&-0.010&-0.257&0.051&-0.167&-0.298&0.275&-0.195&-0.201&0.487\\
&&&Sd&0.026&0.031&0.001&0.031&0.037&0.036&0.038&0.030&0.035&0.036&0.030&0.036&0.034&0.028\\
&Part1 &0.4& Est.&-&1.084&-&-&-0.057&-0.058&-&-&-&-&-&-0.125&-&0.449\\
&&&Sd&-&0.055&-&-&0.084&0.075&-&-&-&-&-&0.082&-&0.036\\
&&0.3& Est.&-&1.104&-&-&-0.118&-0.072&-&-&-&-&-&-0.187&-&0.446\\
&&&Sd&-&0.054&-&-&0.073&0.068&-&-&-&-&-&0.070&-&0.036\\
&&0.2& Est.&-&1.058&-&-&-0.262&-0.084&-&-&-&-&-&-0.254&-&0.441\\
&&&Sd&-&0.046&-&-&0.056&0.054&-&-&-&-&-&0.055&-&0.034\\
&Part 2&0.4& Est.&-0.175&-&0.014&0.106&-&-&-&0.015&-0.270&-0.352&0.239&-0.187&-0.206&-\\
&&&Sd&0.035&-&0.001&0.050&-&-&-&0.047&0.072&0.074&0.045&0.073&0.056&-\\
&&0.3& Est.&-0.175&-&0.014&0.097&-&-&-&0.027&-0.248&-0.368&0.235&-0.292&-0.173&-\\
&&&Sd&0.034&-&0.001&0.048&-&-&-&0.045&0.061&0.062&0.044&0.062&0.053&-\\
&&0.2& Est.&-0.182&-&0.014&0.089&-&-&-&0.036&-0.238&-0.338&0.285&-0.206&-0.215&-\\
&&&Sd&0.031&-&0.001&0.039&-&-&-&0.038&0.046&0.046&0.037&0.047&0.043&-\\
B&&& Est.&-0.109&1.318&0.008&0.176&-0.167&-0.077&-0.189&0.094&0.063&-0.201&0.339&-0.256&-0.242&0.063\\
&&&Sd&0.002&0.004&0.000&0.003&0.010&0.006&0.011&0.003&0.007&0.008&0.003&0.009&0.003&0.002\\

 \multicolumn{18}{l}{$^{*}$ COPD: Chronic obstructive pulmonary; CRF:  Chronic renal failure; EOC:  Exposed to other cases diagnosed as SARS CoV-2. }\\
  \multicolumn{18}{l}{$^{*}$  R and A stand for Random and Adaptive samplings, respectovely, and B is Baseline estimation.}
\end{tabular}
\end{center}
\end{table*}

\begin{table*}[ht]
\caption{Parameter estimation for CoVid data with $d_2=0.05$ and different proportion C2.}
\label{tabr2}\tabcolsep=1pt\fontsize{6}{9}\selectfont
 \vskip 0pt
\par
\begin{center}
\begin{tabular}{ccccc cccc cccc cc ccc}

&&$d_1$&&Gender&  Pneumonia & Age & Diabetes& COPD & Asthma & Immu & Hypertension & Other disease & Card & Obesity & CRF & Smoke & EOC\\
 R&All&0.4& Est.&-0.156&1.094&0.012&0.163&-0.256&-0.173&-0.378&0.038&-0.147&-0.318&0.221&-0.313&-0.101&0.243\\
&&&Sd&0.012&0.027&0.000&0.023&0.081&0.043&0.083&0.021&0.055&0.057&0.019&0.060&0.027&0.013\\
&&0.3& Est.&-0.151&1.086&0.012&0.160&-0.294&-0.156&-0.332&0.040&-0.152&-0.326&0.239&-0.266&-0.133&0.238\\
&&&Sd&0.010&0.021&0.000&0.018&0.060&0.033&0.062&0.016&0.041&0.044&0.015&0.046&0.021&0.010\\
&&0.2& Est.&-0.150&1.078&0.012&0.156&-0.280&-0.144&-0.355&0.037&-0.159&-0.264&0.236&-0.334&-0.186&0.231\\
&&&Sd&0.007&0.014&0.000&0.012&0.040&0.022&0.042&0.011&0.028&0.030&0.010&0.032&0.015&0.007\\
&Part1 &0.4& Est.&-&1.086&-&-&-0.067&-0.199&-&-&-&-&-&-0.341&-&0.256\\
&&&Sd&-&0.043&-&-&0.129&0.067&-&-&-&-&-&0.097&-&0.020\\
&&0.3& Est.&-&1.114&-&-&-0.151&-0.227&-&-&-&-&-&-0.296&-&0.264\\
&&&Sd&-&0.034&-&-&0.103&0.052&-&-&-&-&-&0.074&-&0.015\\
&&0.2& Est.&-&1.096&-&-&-0.296&-0.152&-&-&-&-&-&-0.249&-&0.252\\
&&&Sd&-&0.023&-&-&0.067&0.036&-&-&-&-&-&0.050&-&0.011\\
&Part 2&0.4& Est.&-0.160&-&0.012&0.156&-&-&-&0.022&-0.200&-0.366&0.206&-0.260&-0.105&-\\
&&&Sd&0.018&-&0.001&0.033&-&-&-&0.030&0.082&0.089&0.028&0.089&0.041&-\\
&&0.3& Est.&-0.163&-&0.012&0.176&-&-&-&0.039&-0.167&-0.312&0.210&-0.289&-0.129&-\\
&&&Sd&0.014&-&0.000&0.026&-&-&-&0.024&0.063&0.067&0.022&0.069&0.032&-\\
&&0.2& Est.&-0.158&-&0.012&0.165&-&-&-&0.023&-0.162&-0.306&0.222&-0.293&-0.147&-\\
&&&Sd&0.010&-&0.000&0.018&-&-&-&0.017&0.043&0.046&0.015&0.048&0.022&-\\
A&All&0.4& Est.&-0.170&1.101&0.014&0.094&-0.241&-0.094&-0.418&0.026&-0.260&-0.373&0.239&-0.302&-0.185&0.440\\
&&&Sd&0.034&0.050&0.001&0.048&0.063&0.061&0.065&0.045&0.061&0.063&0.044&0.062&0.053&0.036\\
&&0.3& Est.&-0.178&1.030&0.015&0.079&-0.230&-0.126&-0.331&0.047&-0.237&-0.337&0.253&-0.242&-0.192&0.453\\
&&&Sd&0.032&0.044&0.001&0.043&0.053&0.052&0.053&0.041&0.051&0.051&0.040&0.052&0.047&0.034\\
&&0.2& Est.&-0.166&0.913&0.014&0.117&-0.211&-0.009&-0.232&0.051&-0.170&-0.288&0.284&-0.208&-0.196&0.463\\
&&&Sd&0.027&0.031&0.001&0.031&0.037&0.036&0.038&0.031&0.036&0.036&0.030&0.037&0.034&0.028\\
&Part1 &0.4& Est.&-&1.081&-&-&-0.079&-0.069&-&-&-&-&-&-0.146&-&0.450\\
&&&Sd&-&0.055&-&-&0.083&0.074&-&-&-&-&-&0.077&-&0.036\\
&&0.3& Est.&-&1.094&-&-&-0.143&-0.084&-&-&-&-&-&-0.223&-&0.448\\
&&&Sd&-&0.053&-&-&0.070&0.066&-&-&-&-&-&0.068&-&0.036\\
&&0.2& Est.&-&1.035&-&-&-0.244&-0.101&-&-&-&-&-&-0.271&-&0.439\\
&&&Sd&-&0.045&-&-&0.055&0.053&-&-&-&-&-&0.054&-&0.034\\
&Part 2&0.4& Est.&-0.173&-&0.014&0.108&-&-&-&0.022&-0.279&-0.327&0.235&-0.180&-0.204&-\\
&&&Sd&0.035&-&0.001&0.050&-&-&-&0.046&0.070&0.072&0.045&0.071&0.056&-\\
&&0.3& Est.&-0.173&-&0.014&0.100&-&-&-&0.026&-0.262&-0.362&0.230&-0.303&-0.186&-\\
&&&Sd&0.034&-&0.001&0.047&-&-&-&0.044&0.060&0.061&0.044&0.062&0.052&-\\
&&0.2& Est.&-0.178&-&0.014&0.095&-&-&-&0.054&-0.234&-0.297&0.267&-0.245&-0.220&-\\
&&&Sd&0.031&-&0.001&0.039&-&-&-&0.037&0.045&0.045&0.037&0.046&0.043&-\\

 \multicolumn{17}{l}{$^{*}$ COPD: Chronic obstructive pulmonary; CRF:  Chronic renal failure; EOC:  Exposed to other cases diagnosed as SARS CoV-2.}\\
  \multicolumn{18}{l}{$^{*}$  R and A stand for Random and Adaptive samplings, respectively.}
\end{tabular}
\end{center}
\end{table*}

\section{Discussion and conclusion}
From a statistical perspective, we introduce the idea of the distributed sequential method to the scenarios of federated learning applications without alternating its computing framework. Thus, we can conduct the sequential procedures separately using individual data pools.
The distributed computing feature naturally bypasses latent issues attributed to data communication and data management \citep{lindell2005secure, feigenbaum2001secure, Dwork2011, yan2013,li2018, Carlini2019}. 
The advantages of the sequential estimation  based methods are that they are data driven and have a feature that guarantees the precision of the estimate when sampling is stopped. This feature makes the final ensemble result more stable than other combination methods. 
The adaptive sampling strategy, which is founded on statistical experimental design and information theory allows us to select the most informative samples from a massive data pool, which is a common situation in many data collection scenarios, such as the pandemic data used in our study. Although we only discuss the estimation problem, this idea can be extended to other statistical inferences, which will be examined in furture studies.

\section*{Acknowledgment}

This research is supported in part by research grants from the National Natural Science Foundation of China
(No. 11971457), and 
National Science and Technology Council of Taiwan (No. 111-2118-M-001 -003 -MY2 and No. 108-2118-M-001 -004 -MY3).

\section*{Appendix}
\section*{Proofs of main results}

Throughout the proofs, notation $C$ denotes a generic positive constant, which does not depend on $n$.

\subsection*{A. Properties of sequential estimation for procedure $j$}

When the stopping criterion (8) 
holds for procedure $j$, we record the stopping time and stop sampling.  The confidence ellipsoid for $\vesub{\theta}{0}$ is
\begin{align}
R_{\tilde{N}_j} = \{z\in R^{p_0}: \frac{S_{\tilde{N}_j}}{\tilde{N}_j}\leq \frac{d_1^2}{\mu_{j\tilde{N}_j}}\},\label{Rn-single}
\end{align}
where $S_{\tilde{N}_j}=(\ve z-\tvesub{\theta}{j\tilde{N}_j})^{\top}(\vesub{L}{j}\vess{\Sigma}{j\tilde{N}_{j}}{-1} \vesub{L}{j})^{-1}(\ve z-\tvesub{\theta}{j\tilde{N}_j})$, and $\ve z=(z_1,\cdots, z_{p_0})^{\top}$.
Length of maximum axis of this  ellipsoid is
$$D=2\left(\frac{\tilde{N}_jd_1^2}{\mu_{j\tilde{N}_j}}\right)^{1/2}\lambda^{1/2}_{max}(\tilde{N}_j(\vesub{L}{j}\vess{\Sigma}{j\tilde{N}_{j}}{-1} \vesub{L}{j}))=2d_1
,$$
where $\lambda_{max}(A)$ is the maximum eigenvalue of matrix $A$.
Then we have following Lemmas.

\noindent{\bf Proof of Lemma 1}
Since $\tilde{a}_j^2>0$, for all $j$, are constants, the definition of the stopping time implies that for each $j$, $\tilde{N}_j / \hat{N}$ converges to some constant $\gamma_j>0$. Let $\theta^* = \sum_{j=1}^M w_j \tvesub{\theta}{j\tilde{N}_j}$, where $w_j$, $\sum_{j=1}^M w_j = 1$ are given weights. Then variance of $\theta^*$ is
$$ \sum_{j=1}^M w^2_j {\vesub{L}{j}}\Sigma_{j\tilde{N}_j}^{-1} {\vesub{L}{j}} = \sum_{j=1}^M w^2_j  \tilde{N}_j^{-1} {\vesub{L}{j}}(\Sigma_{j\tilde{N}_j}/\tilde{N}_j)^{-1} {\vesub{L}{j}}.$$
Let
$$ G_N(w_1,\cdots,w_M) = \hat{N}\sum_{j=1}^M w^2_j  \tilde{N}_j^{-1} {\vesub{L}{j}} (\Sigma_{j\tilde{N}_j}/\tilde{N}_j)^{-1} {\vesub{L}{j}},$$
then it follows that as $d_1$ tends to 0,
$$G_N(w_1,\cdots,w_M)\longrightarrow G(w_1,\cdots,w_M)=\sum_{j=1}^M w^2_j \gamma_j^{-1} {\vesub{L}{j}} \Sigma_{j}^{-1} {\vesub{L}{j}}.$$
If we use the same covariates for all data sites, then they have a homogeneous covariance $\Sigma_1=\Sigma_2=\cdots=\Sigma_M$, asymptotically.   We then minimize $G(w_1,\cdots,w_M)$ with respect to $w_1,\cdots,w_M$ subject to $\sum_jw_j=1$, which gives that $w_j=\gamma_j$, $j=1,\cdots,M$.
It follows that $\hve{\theta}$ with the weights $\rho_j$ achieves the minimal covariance, asymptotically.

\noindent LEAMMA C1. {\it {
Assume that the conditions of Theorem 1 
are satisfied, and $\tilde{N}_j$ is as defined in (8). 
Then
\begin{align}
&\lim_{d\rightarrow 0}\frac{d^2\tilde{N}_j}{\tilde{a}_j^2\mu}=1 ~~~~\text{\rm almost surely},  \\
&\lim_{d\rightarrow 0} P(\theta_0 \in R_{\tilde{N}_j})=1-\tilde{\alpha}_j, \\
&\lim_{d\rightarrow 0}\frac{d^2E(\tilde{N}_j)}{\tilde{a}_j^2\mu}=1,
\end{align}
where $\tilde{\alpha}_j$ satisfies $P(\chi_p^2>\tilde{a}_j^2)=\tilde{\alpha}_j$, and $\mu$ is the maximum eigenvalue of matrix $\vesub{L}{j}\Sigma^{-1}\vesub{L}{j}$.
}
}

\noindent Proof. For a single linear regression model, \cite{wangchang13} obtained asymptotic consistency and asymptotic efficiency of the sequential estimation.
Similar to \cite{wangchang13}, proof of this lemma for logistic regression model (14) 
is straightforward and is omitted here.

\subsection*{B. Proof of Theorem 1}

If the stopping criterion defined in (8) 
 holds for the procedure $j$ , then we have used $\tilde{N}_j$ observations, and obtained a confidence set $R_{\tilde{N}_j}$ as defined in (\ref{Rn-single}). 
Via Lemma  C1 
\begin{align}
&\lim_{d_1\rightarrow 0}\frac{d_1^2\tilde{N}_j}{\tilde{a}_j^2\mu}=1 ~~~~\text{\rm almost surely}, \label{conclu1} \\
&\lim_{d_1\rightarrow 0} P(\theta_0 \in R_{\tilde{N}_j})=1-\tilde{\alpha}_j, \label{conclu2}\\
&\lim_{d_1\rightarrow 0}\frac{d_1^2E(\tilde{N}_j)}{\tilde{a}_j^2\mu}=1, \label{conclu3}
\end{align}
where $\tilde{\alpha}_j$ satisfies $P(\chi_p^2>\tilde{a}_j^2)=\tilde{\alpha}_j$. 
%
Equations (\ref{conclu1}) and (\ref{conclu3}) implies that as $d_1\rightarrow 0$,
\begin{align}
&d_1^2\tilde{N}_j\longrightarrow \tilde{a}_j^2 \mu~~~~\text{\rm almost surely},\nonumber \\
&d_1^2E(\tilde{N}_j)\longrightarrow \tilde{a}_j^2 \mu. \nonumber
\end{align}
Because $\tilde{a}^2_1+\tilde{a}^2_2=a^2$,  it follows that
\begin{align}
&d_1^2\hat{N}=d_1^2(\tilde{N}_1+\tilde{N}_2)\longrightarrow (\tilde{a}_1^2 +\tilde{a}_2^2)\mu=a^2\mu~~~~\text{\rm almost surely},\nonumber \\
&d_1^2E(\hat{N})=d_1^2E(\tilde{N}_1+\tilde{N}_2)\longrightarrow (\tilde{a}_1^2 +\tilde{a}_2^2)\mu=a^2\mu, \nonumber
\end{align}
which implies that
\begin{align}
&\lim_{d_1\rightarrow 0}\frac{d_1^2\hat{N}}{a^2\mu}=1 ~~~~\text{\rm almost surely}, \nonumber \\
&\lim_{d_1\rightarrow 0}\frac{d_1^2E(\hat{N})}{a^2\mu}=1. \nonumber
\end{align}

\color{black}{

For simplicity, we suppose that $j=1, 2$, in the following matrix algebra, which csn be easily extended to the case for $j=1, \cdots,  M$.  Matrix $\vesub{L}{j}\vess{\Sigma}{j\tilde{N}_j}{-1}\vesub{L}{j}$ can be used to estimate covariance of $\tvesub{\theta}{j\tilde{N}_j}$.
Since two procedures are independent, we can use
$\rho_1^2\vesub{L}{j}\vess{\Sigma}{1N_1}{-1}\vesub{L}{j}+\rho_2^2\vesub{L}{j}\vess{\Sigma}{2N_2}{-1}\vesub{L}{j}$
to estimate variance  of $\hve{\theta}$.
We already have that $\sqrt{\tilde{N}_j}(\tvesub{\theta}{\tilde{N}_j}-\vesub{\theta}{0})$ has asymptotic normality as $d_1$ tends to 0.
It follows that as $d_1\rightarrow 0$, $\hve{\theta}$ follows an asymptotic normal distribution, and 
\begin{align}
(\hve{\theta}-\vesub{\theta}{0})^\top \left[\rho_1^2\vesub{L}{j}\vess{\Sigma}{1N_1}{-1}\vesub{L}{j}+\rho_2^2\vesub{L}{j}\vess{\Sigma}{2N_2}{-1}\vesub{L}{j}\right]^{-1}(\hve{\theta}-\vesub{\theta}{0})\longrightarrow \chi_{p_0}^2.\label{dist}
\end{align}
By definition of $\mu_{\hat{N}}$, $\mu_{\hat{N}}\rightarrow \mu$ and  ${\hat{N}d_1^2}/{\mu_{\hat{N}}}\rightarrow a^2$ almost surely,  as $d_1\rightarrow 0$.  Thus, (\ref{dist}) implies that
\begin{align}
&\lim_{d_1\rightarrow 0} P(\vesub{\theta}{0} \in R_{\hat{N}})\nonumber\\
&= \lim_{d_1\rightarrow 0} P\left((\hve{\theta}-\vesub{\theta}{0})^\top \left[\rho_1^2\vesub{L}{j}\vess{\Sigma}{1N_1}{-1}\vesub{L}{j}+\rho_2^2\vesub{L}{j}\vess{\Sigma}{2N_2}{-1}\vesub{L}{j}\right]^{-1}(\hve{\theta}-\vesub{\theta}{0})\leq \frac{\hat{N}d_1^2}{\mu_{\hat{N}}}\right)\nonumber\\
&= \lim_{d_1\rightarrow 0} P\left((\hve{\theta}-\vesub{\theta}{0})^\top \left[\rho_1^2\vesub{L}{j}\vess{\Sigma}{1N_1}{-1}\vesub{L}{j}+\rho_2^2\vesub{L}{j}\vess{\Sigma}{2N_2}{-1}\vesub{L}{j}\right]^{-1}(\hve{\theta}-\vesub{\theta}{0})\leq a^2\right)\nonumber\\
&= 1-\alpha.\nonumber
\end{align}
This completes the proof of Theorem 1. 
}

\bibliographystyle{biometrika}
\bibliography{DFS}

\end{document}